\documentclass[10pt,twocolumn,letterpaper]{article}
\usepackage[pagenumbers]{cvpr} % To force page numbers, e.g. for an arXiv version
\usepackage[accsupp]{axessibility} % Improves PDF readability for those with visual impairments.
\usepackage{math}
\usepackage{bm}
\usepackage[dvipsnames]{xcolor}
\usepackage[most]{tcolorbox}
\usepackage{multirow, multicol}
\usepackage{tikz}
\usepackage{rotating}
\usepackage{tablefootnote}
\usepackage{bbm}

\usetikzlibrary{svg.path}
\include{microtype}

\usepackage[dvipsnames]{xcolor}

\definecolor{cvprblue}{rgb}{0.21,0.49,0.74}
\usepackage[pagebackref,breaklinks,colorlinks,citecolor=cvprblue]{hyperref}

\def\paperID{****} % *** Enter the Paper ID here
\def\confName{CVPR}
\def\confYear{2024}

\title{{\ours}: Local Appearance Editing for Neural Radiance Fields}

\author{Lukas Radl \quad {Michael Steiner} \quad {Andreas Kurz} \quad {Markus Steinberger}\\
{\tt\small \{lukas.radl, michael.steiner, andreas.kurz, steinberger\}@icg.tugraz.at} \\
Graz University of Technology \\
}

\begin{document}
\twocolumn[{
\maketitle
\begin{center}
    \centering
    \captionsetup{type=figure}
    \includegraphics[width=\textwidth]{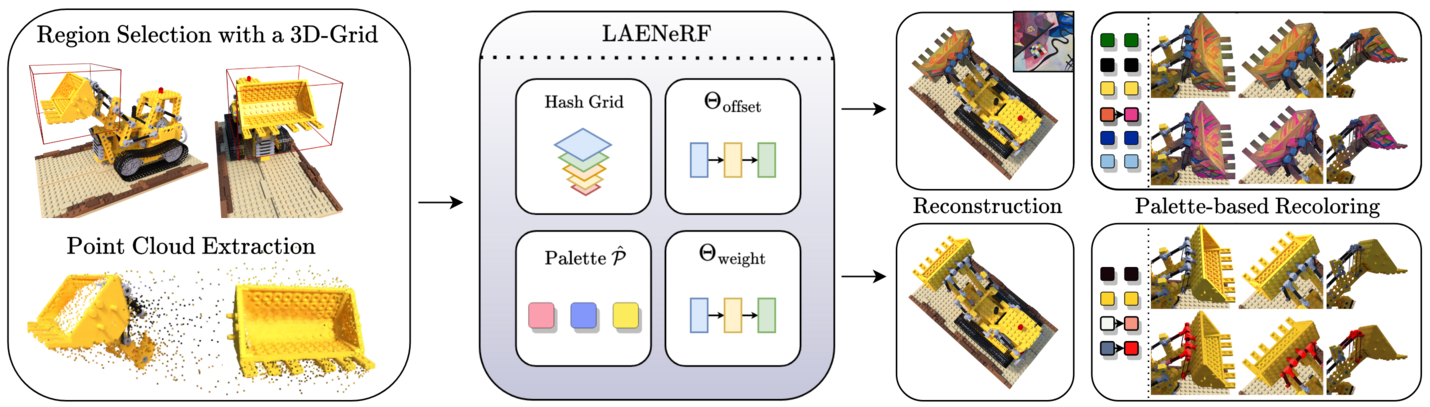}
    \captionof{figure}{
        We propose {\textbf{\ours}}, a method for \textbf{L}ocal \textbf{A}ppearance \textbf{E}diting of Neural Radiance Fields.
        {\ours} enables appearance edits of arbitrary content in 3D scenes while minimizing background artefacts.
        For a specified selection, we learn a mapping from estimated ray termination to output colors via a palette-based formulation, which may be supervised by a style loss.
        In this way, we elegantly combine photorealistic recoloring and non-photorealistic stylization of arbitrary content represented by a radiance field in an interactive framework.
    }
    \label{fig:teaser}
\end{center}%
}]

\begin{abstract}
Due to the omnipresence of Neural Radiance Fields (NeRFs), the interest towards editable implicit 3D representations has surged over the last years.
However, editing implicit or hybrid representations as used for NeRFs is difficult due to the entanglement of appearance and geometry encoded in the model parameters.
Despite these challenges, recent research has shown first promising steps towards photorealistic and non-photorealistic appearance edits.
The main open issues of related work include limited interactivity, a lack of support for local edits and large memory requirements, rendering them less useful in practice.
We address these limitations with {\ours}, a unified framework for photorealistic and non-photorealistic appearance editing of NeRFs.
To tackle local editing, we leverage a voxel grid as starting point for region selection.
We learn a mapping from expected ray terminations to final output color, which can optionally be supervised by a style loss, resulting in a framework which can perform photorealistic and non-photorealistic appearance editing of selected regions.
Relying on a single point per ray for our mapping, we limit memory requirements and enable fast optimization.
To guarantee interactivity, we compose the output color using a set of learned, modifiable base colors, composed with additive layer mixing.selection.
Compared to concurrent work, {\ours} enables recoloring and stylization while keeping processing time low.
Furthermore, we demonstrate that our approach surpasses baseline methods both quantitatively and qualitatively.
\end{abstract}    
\section{Introduction}
\label{sec:intro}
Novel view synthesis has been completely revolutionized by Neural Radiance Fields (NeRFs)~\cite{Mildenhall2020NeRF}.
NeRFs enable high-fidelity reconstruction of a 3D scene from a set of input images and their camera poses, building on differentiable volume rendering.
Recent methods have successfully applied NeRFs to dynamic scenes~\cite{pumarola2021d, zheng2023editablenerf, zielonka2023instant}, large-scale scene reconstruction~\cite{barron2023zip, KurzAdaNeRF, tancik2022blocknerf} and
varying lighting conditions~\cite{chen2022hallucinated, martin2021nerfw}.
Local appearance editing of these learned 3D scene representations remains relatively underexplored.
The implicit representation used in NeRFs in the form of a Multi-Layer Perceptron (MLP) is the main challenge, causing non-local effects when a single parameter is modified.
Distilled feature fields~\cite{kobayashi2022distilledfeaturefields} and per-image 2D masks~\cite{Lee2023ICENeRF} have been suggested to facilitate local edits for NeRFs.
However, both of these methods frequently introduce artefacts in the non-edited regions.
Other editing approaches support recoloring a NeRF by remapping individual colors~\cite{gong2023recolornerf, kuang2022palettenerf, wu2022palettenerf}, try to extract modifiable material quantities for re-rendering~\cite{boss2021nerd, Ye2023IntrinsicNeRF}, or apply style transfer \cite{huang2021learning, zhang2023refnpr}. 
Virtually all approaches in these domains do not support controllable local edits, i.e., they always also introduce global changes, which constrains their viability to the theoretical domain and impedes their applicability in practice. 
At the same time, most approaches struggle with high memory requirements and long compute times, further hampering their use.
Ultimately, no method currently enables simultaneous style transfer and interactive recoloring.

To address the previously discussed limitations, we propose {\ours}, a method for local appearance editing of pre-trained NeRFs.
Choosing NeRFShop~\cite{jambon2023nerfshop} and Instant-NGP (\ngp)~\cite{mueller2022instant} as building blocks, we use a 3-dimensional grid, a subset of {\ngp}'s occupancy grid, as our primitive for selecting scene content.
Due to the region growing procedure inherited from NeRFShop, which relies on a growing queue storing direct neighbors, we can model smooth transitions to content adjacent to our selection, resulting in more visually appealing edits.

Inspired by previous recoloring approaches~\cite{gong2023recolornerf, kuang2022palettenerf}, we introduce a novel NeRF-like module designed to learn a palette-based decomposition of colors within a selected region.
In contrast to previous work, we estimate a per-ray termination point resulting in a point cloud which represents the editable region.
This design decision reduces memory requirements and increases performance drastically.
We feed these points into our neural {\ours} module, which learns a palette-based decomposition by jointly optimizing two MLPs and a set of base colors to reconstruct the selected region (see \cref{fig:teaser}).
As {\ours} learns a function in 3D space, we can implicitly ensure multi-view consistency and prune outliers.
The learned set of colors may be modified after optimization to enable interactive recoloring.

By providing a style loss during reconstruction, {\ours} can stylize the selected region, while keeping its recoloring abilities and processing time low.
We propose several novel losses to generate high-fidelity results while respecting the learned 3D geometry and extracting an intuitive color decomposition.
Finally, we generate a modified training dataset by blending our edited region with the original training dataset and fine-tune the pre-trained NeRF.
Our experiments demonstrate that {\ours} is not only the first interactive approach for NeRF appearance editing, but also qualitatively and quantitatively outperforms previous methods for local recoloring and stylization. 

\textbf{In summary}, we make the following contributions:
\begin{enumerate}
    \item[(1)] We combine photorealistic and non-photorealistic appearance edits for NeRFs into a unified framework.
    \item[(2)] We propose the first interactive approach for local, recolorable stylization of arbitrary regions in NeRFs.
    \item[(3)] We propose a new architecture and novel regularizers for efficient, geometry-aware 3D stylization. 
\end{enumerate}

\section{Related Work}
\label{sec:related}

\paragraph{Photorealistic Appearance Editing}
\label{sec:related:photor}
i.e. recoloring for NeRFs, modifies the underlying material colors, without changing textures or lighting.
For this task, several methods~\cite{gong2023recolornerf, kuang2022palettenerf, Kenji2022Posterization, wu2022palettenerf} learn a decomposition into a set of base colors with barycentric weights and per-pixel offsets with additive layer mixing~\cite{tan2018palette, tan2016decomposing}.
This color palette can be modified interactively during inference.
Orthogonally, several approaches recover the material properties directly~\cite{boss2021nerd, tang2022nerf2mesh, verbin2022refnerf, verbin2023eclipse, Ye2023IntrinsicNeRF, zhang2021nerfactor}.
To further enable local recoloring, feature fields can be jointly optimized during training~\cite{kobayashi2022distilledfeaturefields}.
{\pnf}~\cite{kuang2022palettenerf} incorporates this idea, allowing end users to guide recoloring given the selection of a single reference point.
However, this approach limits the end user when performing local edits and is prone to introducing artefacts.
{\icenerf}~\cite{Lee2023ICENeRF} performs local recoloring by modifying the most significant weights in the color MLP of a trained NeRF, given per-image annotations of foreground and background.
This approach works well for bounded or forward-facing scenes, but struggles for unbounded, $360^\circ$ captures due to the large 3D space.
In comparison to all previous methods, our approach ensures that edits remain local and provides an intuitive interface to artists.

\paragraph{Non-Photorealistic Appearance Editing}
\label{sec:related:nonphotor}
for NeRFs modifies textures as well as material properties. 
Leveraging image style transfer~\cite{huang2017arbitrary, gatys2016styletransfer, gatys2017controlling}, recent methods apply perceptual losses~\cite{gatys2016styletransfer, johnson2016perceptual} to radiance fields.
Among these, several propose separate modules for color and style~\cite{fan2022unified, huang2021learning, huang2022stylizednerf, Pang2023LocallyStylized} or progressively stylize a trained NeRF scene~\cite{nguyen2022snerf, wang2023nerfart, zhang2022arf, zhang2023lipschitznet}.
Another line of works utilizes two separate NeRFs, one for reconstruction and one for style~\cite{bao2023sine, gordon2023blendednerf, wang2023inpaintnerf360}.
Recent methods have also investigated modifiable stylization:
Pang~\etal~\cite{Pang2023LocallyStylized} can generate different versions of the same style by utilizing a hash grid~\cite{mueller2022instant} with modifiable hash coefficients.
StyleRF~\cite{liu2023stylerf} utilizes a feature field for global, zero-shot stylization.
{\refnpr}~\cite{zhang2023refnpr} faithfully propagates the style of a reference image to a pre-trained NeRF.
Our approach requires significantly less memory and compute resources, leads to higher multi-view-consistency, allows to apply style transfer only locally, and enables recoloring of the stylized radiance field within an interactive framework.
\section{Preliminaries}
\label{sec:prelim}
In this section, we revisit volumetric rendering with NeRFs and outline our procedure for region selection.

\paragraph{Neural Radiance Fields.}
NeRFs~\cite{Mildenhall2020NeRF} learn a function
\begin{equation}
    \nerf: \R^5 \rightarrow \R^4,\ (\vec{p}, \vec{d}) \mapsto (\vec{c}, \sigma),
\end{equation}
where $\vec{p} = (x,y,z)$ and $\vec{d} = (\theta, \phi)$ denote the sample position and viewing direction, $\vec{c} \in [0,1]^3$ denotes the predicted output color and $\sigma \in \R_+$ denotes the predicted volumetric density. 
For each pixel, positions $\vec{r}(t) = \vec{o} + t\vec{d}$ along a ray from the camera position $\vec o$ in the direction $\vec d$ are sampled. 
At $N$ points $t_i: 0 < i \leq N$ along $\vec r$ between the near and far plane, the colors and density are evaluated and composed using volumetric rendering:
\begin{equation}
    \hat{\mathcal{C}}(\vec{r}) = \sum_{i=1}^{N} T_i (1 - \exp({-\sigma_i \delta_i}))\vec{c}_i,
\end{equation}
where $\delta_i = t_{i+1}-t_i$ and the transmittance $T_i$ is given as
\begin{equation}
    T_i = \exp{\left(- \sum_{j=1}^{i-1} {\sigma_j \delta_j}\right)}.
\end{equation}
Through this fully differentiable pipeline, $\nerf$ can be optimized with $\sum_{\vec r \in \mathcal{R}} \|\hat{\mathcal{C}}(\vec{r}) - {\mathcal{C}}(\vec{r})\|_2^2$ for a set of rays $\mathcal{R}$,
where ${\mathcal{C}}(\vec{r})$ denotes the ground truth color.
With a trained $\nerf$, we can compute the estimated depth $\depth$ using
\begin{equation}
    \depth = \sum_{i=1}^{N} T_i (1 - \exp({-\sigma_i \delta_i})) t_{i+1}\label{eq:prelim:depth},
\end{equation}
which we use to compute the estimated ray termination
\begin{equation}
    \xterm = \vec o + \depth \vec d. \label{eq:prelim:xterm}
\end{equation}

\paragraph{Region Selection with NeRFShop.}
NeRFShop~\cite{jambon2023nerfshop} leverages a voxel grid, akin to the occupancy grid $\occgrid$ of {\ngp}~\cite{mueller2022instant}, to select arbitrary content in {\ngp}-based radiance fields.
The occupancy grid of {\ngp} is an acceleration structure, discretizing the bounded domain into uniformly sized voxels, with a value of 0 or 1 describing the expected density within this voxel.
We utilize this concept with our edit grid $\editgrid$, following NeRFShop~\cite{jambon2023nerfshop}.

To facilitate region selection, the user scribbles on the projection of the 3D scene on the screen.
Subsequently, for each selected pixel, a ray is cast and the estimated ray termination $\xterm$ is computed with \cref{eq:prelim:xterm}.
Next, we map each $\xterm$ to the nearest voxel in our edit grid $\editgrid$ and set the corresponding bit in the underlying bitfield.
For intuitive selection, we support user-controlled region growing from the selected voxels, by adding the neighboring voxels to a growing queue $\growingqueue$.
During region growing, we add the current voxel to $\editgrid$ and its neighbors to $\growingqueue$ if the $\occgrid$ is set.
Through this workflow, we offer an intuitive method for content selection within {\ngp}'s hybrid representation.
\begin{figure*}[t!]
    \centering
    \includegraphics[width=.95\linewidth]{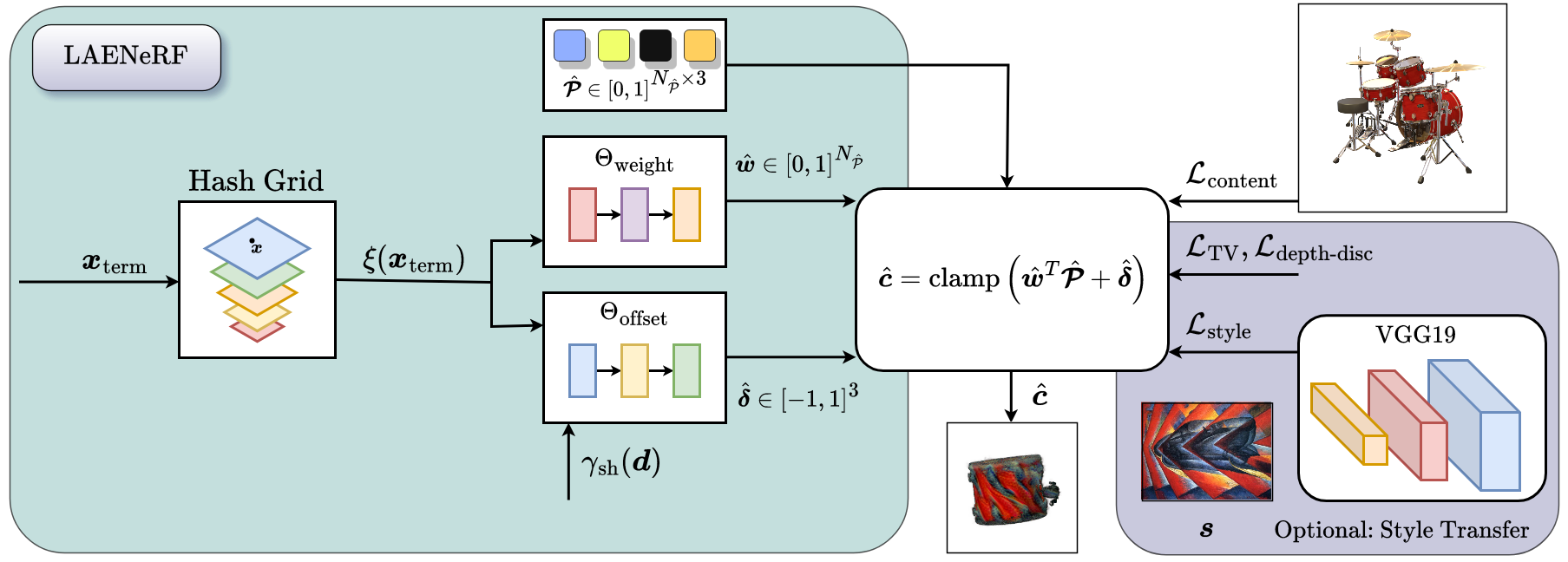}
    \caption{Overview of {\ours}:
    Given the estimated ray terminations $\xterm$ for a region specified by $\editgrid$, we learn a mapping from positions to weights $\what$ and offsets $\dhat$.
    We compose the color $\cpred$ using these latent outputs and a learnable color palette $\palet$ and supervise with a content loss $\lcontent$ and optional style losses $\lstyle, \ltv, \ldepthdisc$ to obtain a unified approach which supports recoloring and stylization.}
    \label{fig:method:ours:arch}
\end{figure*}

\section{{\ours}}
\label{sec:method:ours}
Our key insight is that we can reduce computational requirements by optimizing a lightweight, NeRF-like network given 3D positions $\xterm$, which represents a local, edited region.
Starting from $\editgrid$ defined by the user, we extract $\xterm$ for all training views, optimize LAENeRF and transfer the changes to $\nerf$ with an efficient distillation step.
We present our network architecture in \cref{fig:method:ours:arch}.
$\xterm$ is featurized using a trainable multi-resolution hash grid~\cite{mueller2022instant}.
The encoded input $\xi(\xterm)$ is subsequently passed to shallow MLPs to predict per-ray barycentric weights $\what \in [0,1]^{\numpalet}$ and view-dependent offsets $\dhat \in [-1,1]^3$, where $\numpalet$ denotes the size of the learnable color palette $\palet \in \R^{\numpalet \times 3}$.
Finally, we compose our output color as
\begin{equation}
    \cpred = \text{clamp}\left( \what^T \palet + \dhat \right). \label{eq:method:colorpred}
\end{equation}
We featurize $\vec d$ using a spherical harmonics encoding~\cite{fridovich2022plenoxels, verbin2022refnerf} and use this as an additional input to our offset network.

\paragraph{Input: Estimated Ray Termination.}
As can be seen in \cref{fig:method:ours:arch},\ {\ours} learns a mapping from $\xterm$ to estimated output color $\cpred$.
However, we only require $\xterm$ of rays $\vec r$ which intersect $\editgrid$.
To correctly handle occlusions and mitigate errors due to the edit grid resolution, we require the alpha accumulated inside the edit grid $T_{\alpha}$ to be larger than $\teditgrid=0.5$, where we define $T_{\alpha}$ as
\begin{equation}
    T_{\alpha} = \sum_{i=1}^{N} \mathbbm{1}(\vec o + t_i \vec d \in \editgrid)\ T_i (1 - \exp({-\sigma_i \delta_i})) \label{eq:accum_alpha},
\end{equation}
with $\mathbbm{1}(\cdot)$ denoting the indicator function.
To obtain accurate depth estimates $\depth$, requiring the full accumulation of alpha, we simultaneously raymarch through both $\editgrid$ and the occupancy grid $\occgrid$, computing $\depth$ within the occupancy grid.

\paragraph{Style and Content Losses.}
As is common in image style transfer~\cite{gatys2016styletransfer, johnson2016perceptual}, we use separate losses for content and style:
\begin{align}
    \lcontent &= \|\cpred - \mathcal{C}(\vec{r}) \|_2^2, \label{eq:lcontent}\\
    \lstyle &= \lambdastyle \|\vec{G}({\styleimg}) - \vec{G}({\cpred})\|_2^2 \label{eq:lstyle},
\end{align}
where $\styleimg$ denotes an arbitrary style image and $\vec{G}(\cdot)$ denotes the gram matrix~\cite{gatys2016styletransfer} of a feature extracted from a semantic encoder, \eg VGG19~\cite{Simonyan2014vgg}.
Crucially, we can perform photorealistic recoloring by setting $\lambdastyle=0$, where {\ours} learns to reconstruct the selected region due to~\cref{eq:lcontent}.

\paragraph{Geometry-preserving Losses for Stylization.}
We notice that when we perform non-photorealistic stylization of 3D regions using only $\lcontent$ and $\lstyle$, small structures are often eliminated in favor of a consistent stylization.
In addition, as imperfect geometry reconstruction from our pre-trained $\nerf$ leads to noise in $\xterm$, we need additional regularization to encourage smooth, low-noise outputs.
To facilitate detailed, geometry-aware stylization, we introduce two novel losses, which condition {\ours} on the estimated geometry of $\nerf$. 

First, we encourage {\ours} to limit noise in regions without depth discontinuities using a depth guidance image:
\begin{equation}
    (\depthguide)_{i, j} = \left(| {\depth}_{i, j+1} - {\depth}_{i, j} |,\ | {\depth}_{i+1, j} - {\depth}_{i, j} |\right)\label{eq:method:depthguide}.
\end{equation}
Then, we use this guidance image to restrict a total variation loss to regions without depth discontinuities.
To this end, we introduce our novel, depth-guided total variation loss as
\begin{equation}
    \ltv = \lambdatv \|\nabla \cpred \cdot (1 - \depthguide)\|_2^2,
\end{equation}
where $\nabla \cpred$ denotes the image gradients of $\cpred$ in $x/y$-direction, i.e. $(\nabla \cpred)_{i,j} = \|\cpred_{i+1,j} - \cpred_{i,j}\|_2^2 + \|\cpred_{i,j+1} - \cpred_{i,j}\|_2^2$.
This loss term remedies noisy prediction for $\xterm$, but does not preserve fine, geometric structures sufficiently.
Hence, we introduce another loss, which is minimized when image gradients are placed in regions of depth discontinuities:
\begin{equation}
    \ldepthdisc = -\lambdadepthdisc \|\nabla \cpred \cdot \depthguide\|_2^2.
\end{equation}

\paragraph{Palette Regularization.}
As we learn a palette-based decomposition of output colors given in \cref{eq:method:colorpred}, we require carefully designed regularization to ensure an intuitive color decomposition.
We introduce a weight loss to encourage sparse per-pixel predictions:
\begin{equation}
    \lweight = \lambdaweight \left( 1 - \left\| \what \right\|_\infty \right). \label{eq:method:sparseweights}
\end{equation}
To prevent extreme solutions with a high-frequency offset function, we use the offset loss from Aksoy~\etal~\cite{aksoy2017unmixing}:
\begin{equation}
     \loffset = \lambdaoffset \| \dhat \|_2^2. \label{eq:method:sparseoffsets}
\end{equation}
Finally, we regularize $\palet$ to guarantee valid colors in RGB-space, i.e. ${\palet}_{i,j} \in [0,1]$, using 
\begin{equation}
     \lpalet = \left\| \lfloor \palet \rfloor \cdot \palet \right\|_2^2. \label{eq:method:validpalet}
\end{equation}

\paragraph{Distillation of the Appearance Edits.}
To transfer the local changes encoded in {\ours} to the pre-trained radiance field, we require an additional fine-tuning step.
First, we obtain a modified training dataset, where we compose the target color as $\Talpha \cpred + (1-\Talpha)\mathcal{C}(\vec{r})$.
Thus, rays which did not intersect $\editgrid$ use the ground truth color $\mathcal{C}(\vec{r})$ during distillation, which effectively mitigates background artefacts. 

\begin{figure}[h!]
    \centering
    \includegraphics[width=\linewidth]{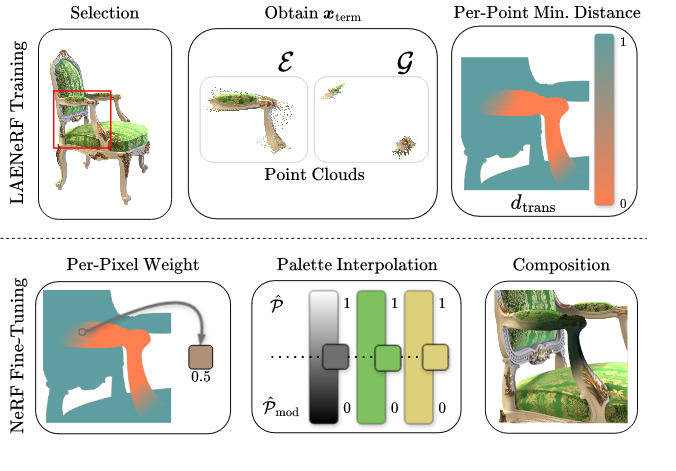}
    \caption{Illustration of our proposed distance-based palette interpolation scheme:
    We calculate distance weights $\dtrans$ based on the per-point distance from the edit grid $\editgrid$ to the growing grid $\growgrid$.
    When constructing the modified training dataset, we interpolate between learned palette $\palet$ and modified palette $\palet_{\text{mod}}$ using $\dtrans$.}
    \label{fig:smooth_trans}
\end{figure}

\paragraph{Modelling Smooth Transitions.}
As our model operates on a user-defined region of interest and utilizes blending to construct a new dataset, recoloring or stylization produces sharp discontinuities on the boundary of $\editgrid$.
While this behaviour is desirable when cells adjacent to $\editgrid$ are not occupied, it may lead to undesirable results otherwise.
As the growing queue $\growingqueue$ stores adjacency information, we can use it to model smooth color transitions for more visually pleasing results.
First, we construct a new grid $\growgrid$ from our growing queue $\growingqueue$.
Then, we raymarch using $\growgrid$ and $\occgrid$ to obtain the estimated ray terminations for each ray intersecting $\growgrid$, resulting in another point cloud, which we denote as $\mathcal{Z}$.
For each $\xterm$ obtained from raymarching through $\editgrid$, we compute the minimum distance to $\mathcal{Z}$:
\begin{equation}
    d_{\text{min}} = \mini{\vec y \in  \mathcal{Z}} \|\xterm - \vec y\|_2.
\end{equation}
We construct per-ray transition weights $\dtrans \in [0,1]$ with
\begin{equation}
     \dtrans = 1 - \frac{\text{min}(d_{\text{min}}, \growdist)}{\growdist},
\end{equation}
where $\growdist$ is a hyperparameter controlling the size of the transition, which we usually set to $1\times 10^{-2}$, depending on the scale of the scene.
As can be seen in \cref{fig:smooth_trans}, we interpolate the original color palette $\palet$ and the user-modified color palette $\palet_{\text{mod}}$ based on $\dtrans$ to get a realistic color transition.
For stylization, we additionally introduce a separate loss, which adds to the content loss when $\dtrans \in (0, 1]$:
\begin{equation}
    \lsmooth = \lambdasmooth \|(\cpred - \vec c)^2 \cdot \dtrans\|_1.
\end{equation}

\paragraph{Implementation Details.}
\label{sec:method:ours:optimization}
We build our approach on \textit{torch-ngp}~\cite{torch-ngp} and use a resolution of $128^3$ for our edit grid $\editgrid$, following iNGP~\cite{mueller2022instant}.
We train {\ours} for $1 \times 10^{5}$ iterations with previews available after $\sim 20$s, and then distill to $\nerf$ for $7 \times 10^{4}$ iterations, which takes no longer than 5 minutes in total on an NVIDIA RTX 4090.
For {\ours}, we use a learning rate of $1 \times 10^{-3}$ for all components, except for $\palet$, where we use $1 \times 10^{-2}$.
We initialize $\numpalet=8$, as the use of $\lsmooth$ requires a large number of base colors.
$1.5 \times 10^{3}$ iterations before training is finished, we remove color palettes which do not contribute significantly.
For our style loss, we use a VGG19 backbone~\cite{Simonyan2014vgg} and utilize features from \texttt{conv5, conv6, conv7} for computing $\lstyle$.
When performing stylization, we use $\lambdastyle = 1.3 \times 10^{2}, \lambdatv = 1 \times 10^{-4}, \lambdadepthdisc = 5 \times 10^{-4}, \lambdaweight = 1 \times 10^{-7}, \lambdaoffset = 5 \times 10^{-5}, \lambdasmooth=1 \times 10^{-3}$.
If we want to perform photorealistic recoloring, we set $\lambdastyle = \lambdatv = \lambdadepthdisc = 0$.
\section{Experiments}
\label{sec:experiments}
We use {\ours} to perform local recoloring and local, recolorable stylization.

\paragraph{Datasets.}
We use three well-established datasets for novel view synthesis in our evaluation.
\textbf{NeRF-Synthetic}~\cite{Mildenhall2020NeRF} is a dataset consisting of synthetic objects with complex geometry and non-Lambertian materials.
This dataset contains $360^{\circ}$ captures in a bounded domain with a transparent background.
\textbf{LLFF}~\cite{mildenhall2019llff} is a dataset consisting of forward-facing captures of real-world scenes in high resolution.
The \textbf{\threesixty} dataset~\cite{Barron2022MipNeRF360} contains $360^{\circ}$ captures of unbounded indoor and outdoor scenes.
This provides a challenging scenario for local appearance editing methods due to the large number of distinct objects in the scene and the large 3D space.
For LLFF and {\threesixty}, we follow related work~\cite{Lee2023ICENeRF, kuang2022palettenerf} and downsample the images by a factor of $4\times$.

\subsection{Photorealistic Appearance Editing}
\label{sec:experiments:photor}
For the recoloring task, we compare our method with {\pnf}~\cite{kuang2022palettenerf} and {\icenerf}~\cite{Lee2023ICENeRF}.
{\pnf} predicts features distilled from an LSeg segmentation model~\cite{kobayashi2022distilledfeaturefields, li2022languagedriven} to enable local editing.
In contrast, {\icenerf} uses user-guided annotations to recolor a selected region given a target color.

\paragraph{Quantitative Evaluation.}
For the quantitative evaluation, we follow {\icenerf}~\cite{Lee2023ICENeRF} and measure the Mean Squared Error (MSE) in the background of the selected region before and after recoloring.
In \cref{tab:pr:quantitative:mse}, we present this metric for three scenes of the LLFF dataset~\cite{mildenhall2019llff}.
The foreground region is described by a mask, which was provided to us by the authors of {\icenerf}.
We perform 7 different recolorings per scene and compare against {\pnf} with and without semantic guidance, whereas we also include the numbers from ICE-NeRF\footnote{{\icenerf}s' implementation is not yet publicly available.} to facilitate cross-method comparisons.
{\ours} outperforms previous methods for this metric, reducing error rates by $59\%$ compared to {\pnf} with semantic guidance.
Additionally, we provide the same metric for the indoor scenes of the {\threesixty} dataset~\cite{Barron2022MipNeRF360} in \cref{tab:pr:quantitative:mse:360}, where we report average results for 7 recolorings per scene.
We use Segment Anything~\cite{kirillov2023segmentanything} to obtain masks for a subset of test set views and compare against {\pnf}~\cite{kuang2022palettenerf}.
We measure MSE compared to the ground truth test images and use the same hyperparameters as {\pnf}, which also builds on torch-ngp~\cite{torch-ngp}.
As can be seen, {\ours} after recoloring outperforms non-recolored {\pnf}, which we attribute to {\pnf}'s concurrent scene reconstruction and palette-based decomposition.
Due to the fairly accurate geometry for the $360^\circ$ captures, our approach introduces very few background artefacts during recoloring.

\begin{table}[t!]
  \centering
  \small
    \setlength{\tabcolsep}{5pt}
    \begin{tabular}{@{}lllllll@{}}\toprule
    Scene       & \multicolumn{2}{c}{{Results from~\cite{Lee2023ICENeRF}}} && \multicolumn{3}{c}{{Our Recolorings}} \\
    \cmidrule{2-3} \cmidrule{5-7}
                & {\scriptsize PNF} & {\scriptsize \icenerf}  &&{\scriptsize PNF} & {\scriptsize PNF} & {\scriptsize \ours}\\[-.9ex]
                &         &          &&&{\scriptsize (Semantic)} &\\[-.7ex]
    \midrule
    \emph{Horns}       & 0.0818 & 0.0213 && 0.0195 & 0.0028 & \textbf{0.0010} \\
    \emph{Fortress}    & 0.0013 & 0.0010 && 0.0011 & \textbf{0.0002} & \textbf{0.0002} \\
    \emph{Flower}      & \textbf{0.0003} & \textbf{0.0003} && 0.0076 & 0.0022 & {0.0007} \\
    \midrule
    Average     & 0.0277 & 0.0075 && 0.0094 & 0.0017 & \textbf{0.0007} \\
    \bottomrule
    \end{tabular}
  \caption{
  MSE ($\downarrow$) in the background with respect to the unmodified images for our method, {\icenerf}~\cite{Lee2023ICENeRF} and {\pnf}~(PNF)~\cite{kuang2022palettenerf} for the LLFF dataset~\cite{mildenhall2019llff}.
  Note that the recolorings from~\cite{Lee2023ICENeRF} are different to ours.
  }
  \label{tab:pr:quantitative:mse}
\end{table}

\begin{table}[t!]
  \small
  \centering
    \setlength{\tabcolsep}{5pt}
    \begin{tabular}{@{}lllllll@{}}\toprule
    Scene & \multicolumn{3}{c}{{\pnf}} && \multicolumn{2}{c}{{\ours}} \\
    \cmidrule{2-4} \cmidrule{6-7}
            & Trained & Recolor & Recolor && Trained & Recolor\\[-.9ex]
            &         &         & {\scriptsize (Semantic)} &&&\\[-.7ex]
    \midrule
    \emph{Bonsai}  & 0.0015 & 0.0036 & 0.0016 && 0.0011 & \textbf{0.0011} \\
    \emph{Kitchen} & 0.0024 & 0.0125 & 0.0027 && 0.0021 & \textbf{0.0022} \\
    \emph{Room}    & 0.0016 & 0.0216 & 0.0056 && 0.0015 & \textbf{0.0015} \\
    \midrule
    Average & 0.0018 & 0.0124 & 0.0033 && 0.0016 & \textbf{0.0016} \\
    \bottomrule
    \end{tabular}
  \caption{
  MSE ($\downarrow$) in the background with respect to the test images for our method and {\pnf}~\cite{kuang2022palettenerf} for the {\threesixty} dataset~\cite{Barron2022MipNeRF360}.
  {\ours} exhibits lower error rates compared to {\pnf}, even when comparing trained with recolored.}
  \label{tab:pr:quantitative:mse:360}
\end{table}

\begin{figure}
    \centering
    \newcommand{\rulesep}{\unskip\ \vrule\ }
     \begin{minipage}[t]{0.2375\linewidth}
        \centering
        \footnotesize
        \text{Ground Truth}
    \end{minipage}
    \begin{minipage}[t]{0.2375\linewidth}
        \centering
        \footnotesize
        \text{{\ours}}
    \end{minipage}
    \begin{minipage}[t]{0.2375\linewidth}
        \centering
        {\footnotesize
        \text{{\icenerf}~\cite{Lee2023ICENeRF}}}\ 
    \end{minipage}
    \begin{minipage}[t]{0.2375\linewidth}
        \centering
        {\footnotesize
        \text{{\pnf}~\cite{kuang2022palettenerf}}} \\[-1ex]
        {\scriptsize
        \text{(Semantic)}}
    \end{minipage} \\
    \includegraphics[width=.98\linewidth]{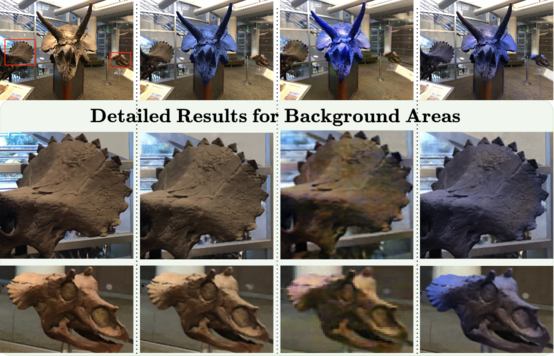}
    \caption{Qualitative comparison of our method with related work on the \emph{Horns} scene of the LLFF dataset~\cite{mildenhall2019llff}. 
    {\ours} introduces far fewer artefacts compared to previous methods.}
    \label{fig:pr:qualitative:regions}
\end{figure}

\begin{figure}[h!]
    \centering
    \newcommand{\rulesep}{\unskip\ \vrule\ }
    \begin{minipage}[t]{0.35\linewidth}
        \centering
        \small
        \text{Ground Truth}
    \end{minipage}
    \begin{minipage}[t]{0.3\linewidth}
        \centering
        \small
        \text{\ours}
    \end{minipage}
    \begin{minipage}[t]{0.3\linewidth}
        \centering
        {\small
        \text{${\pnf}$~\cite{kuang2022palettenerf}} \\[-1ex]
        \scriptsize
        \text{(Semantic)}}
    \end{minipage} \\
        \begin{sideways}
        \hspace{1.5em}
        \footnotesize
        \emph{Bonsai}
    \end{sideways}
    \includegraphics[width=.313\linewidth]{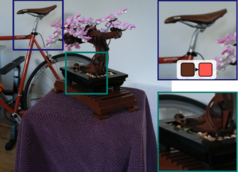}
    \includegraphics[width=.313\linewidth]{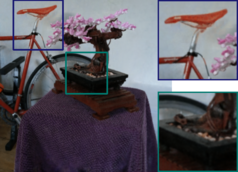}
    \includegraphics[width=.313\linewidth]{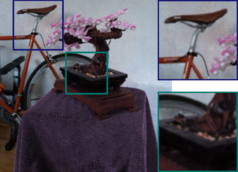} \\
    \begin{sideways}
        \hspace{1.5em}
        \footnotesize
        \emph{Room}
    \end{sideways}
    \includegraphics[width=.313\linewidth]{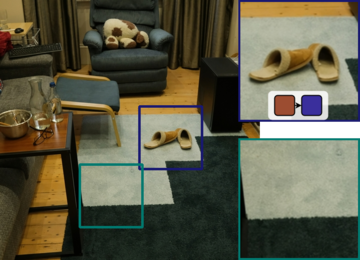}
    \includegraphics[width=.313\linewidth]{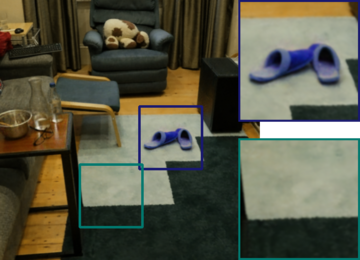}
    \includegraphics[width=.313\linewidth]{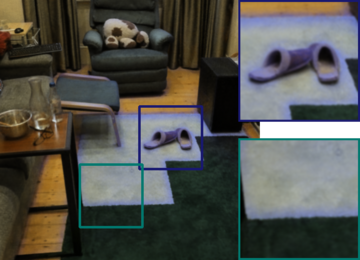}
    \caption{Qualitative comparison of our method to {\pnf} on the {\threesixty} dataset~\cite{Barron2022MipNeRF360} for small-scale edits.
    The top-row detailed view shows the selected region for recoloring, whereas the bottom-row view shows a background region.
    Our method introduces fewer errors in the background whilst recoloring the selected object faithfully.}
    \label{fig:pr:qualitative:360}
\end{figure}

\def\greencheck{\tikz\fill[scale=0.4, color=green](0,.35) -- (.25,0) -- (.8,.6) -- (.25,.15) -- cycle;}
\begin{figure}
    \centering
    \begin{minipage}[t]{0.316\linewidth}
            \centering
            \small
            \text{Multiple Edits}
        \end{minipage}
        \begin{minipage}[t]{0.316\linewidth}
            \centering
            \small
            \text{Smooth Transitions}
        \end{minipage}
            \begin{minipage}[t]{0.316\linewidth}
            \centering
            \small
            \text{Detailed Local Edits}
        \end{minipage}
    \includegraphics[width=.316\linewidth]{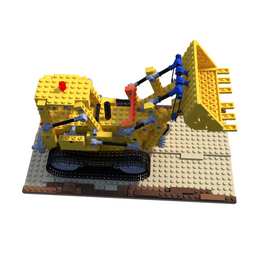}
    \includegraphics[width=.316\linewidth]{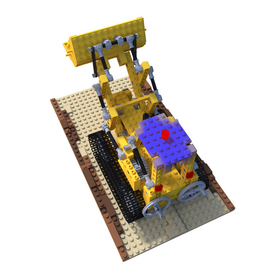}
    \includegraphics[width=.316\linewidth]{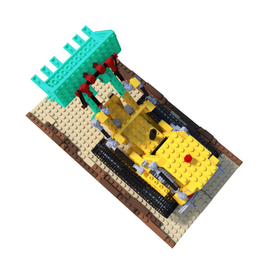} \\
    \begin{minipage}[t]{0.316\linewidth}
            \centering
            \includegraphics[height=.2\linewidth]{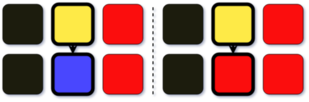}
    \end{minipage}
    \begin{minipage}[t]{0.316\linewidth}
            \centering
            \includegraphics[height=.2\linewidth]{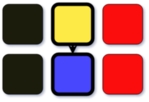}
    \end{minipage}
    \begin{minipage}[t]{0.316\linewidth}
            \centering
            \includegraphics[height=.2\linewidth]{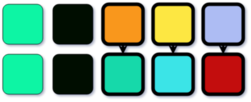}
    \end{minipage}
    \begin{tcolorbox}[colback=MidnightBlue!10,%gray background
                  colframe=MidnightBlue!50,% black frame colour
                  width=.96\linewidth,% Use 5cm total width,
                  boxrule=1pt,
                  boxsep=1mm, left=-1mm, right=-1mm, top=0mm, bottom=0mm,
                  enlarge top by=-2.5mm, enlarge bottom by=-3mm,
                 ]
                 \centering
        \textbf{Zoomed-In Results}\\
        \vspace{0.5mm}
        \includegraphics[width=.158\linewidth]{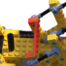}
        \includegraphics[width=.158\linewidth]{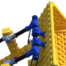} 
        \includegraphics[width=.158\linewidth]{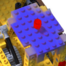}
        \includegraphics[width=.158\linewidth]{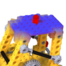}   
        \includegraphics[width=.158\linewidth]{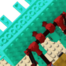}
        \includegraphics[width=.158\linewidth]{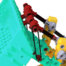} \\
    \end{tcolorbox}
    \caption{
    Demonstration of our editing capabilities: {\ours} can perform arbitrary edits on any local region with smooth transitions.}
    \label{fig:pr:qualitative:unique}
\end{figure}

\paragraph{Qualitative Evaluation.}
In \cref{fig:pr:qualitative:regions}, we compare our method to {\icenerf}~\cite{Lee2023ICENeRF} and {\pnf}~\cite{kuang2022palettenerf}.
As can be seen, our method introduces the fewest artefacts due to recoloring.
To facilitate cross-method comparisons, we choose the same example as {\icenerf} and include the results from their publication.
For the {\threesixty} dataset~\cite{Barron2022MipNeRF360}, we compare against {\pnf} in \cref{fig:pr:qualitative:360}.
{\pnf} either introduces significant artefacts (see \emph{Room}) or is unable to capture small regions with their semantic features (see \emph{Bonsai}). 
As mentioned in their publication, {\icenerf} struggles with local edits for this dataset, frequently introducing artefacts in undesired regions.

In \cref{fig:pr:qualitative:unique}, we show some recoloring results only possible with our method on the synthetic \emph{Lego} scene.
{\pnf}'s semantic features do not permit any of the shown edits.
While {\icenerf} can perform multiple color edits, their approach recolors based on a single target color and thus fails for edits where multiple palette changes are required.
{\ours} is the only method which can model smooth transitions between original scene content and the recolored region.

% ----
% Non-Photorealistic Appearance Editing
% ----

\subsection{Non-Photorealistic Appearance Editing}
\label{sec:experiments:nonphotor}
For style transfer, we compare our method to {\refnpr}~\cite{zhang2023refnpr}, which stylizes a scene based on one or a few reference images.
As {\refnpr} is not specifically designed for local stylization, we create locally stylized reference images by selecting three training dataset images, applying AdaIN~\cite{huang2017arbitrary} for stylization using a style image $\styleimg$, and using {\ours}'s blending weights to generate three reference images. 
Additional details are available in the supplementary material.
\begin{figure}[ht!]
    \centering
    \begin{minipage}[t]{0.316\linewidth}
        \centering
        \text{Ground Truth}
    \end{minipage}
    \begin{minipage}[t]{0.316\linewidth}
        \centering
        \text{LAENeRF}
    \end{minipage}
    \begin{minipage}[t]{0.316\linewidth}
        \centering
        \text{{\refnpr}~\cite{zhang2023refnpr}}
    \end{minipage}\\
    \begin{sideways}
        \hspace{1.5em}
        \footnotesize
        \emph{Drums}
    \end{sideways}
    \includegraphics[width=.3211\linewidth]{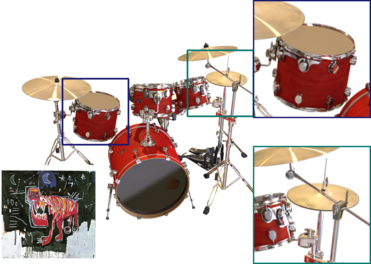}
    \includegraphics[width=.3069\linewidth]{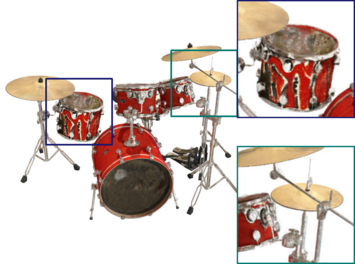}
    \includegraphics[width=.3069\linewidth]{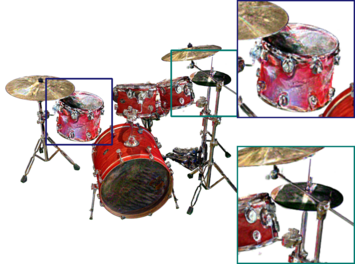} \\
    \begin{sideways}
        \hspace{1.5em}
        \footnotesize
        \emph{Horns}
    \end{sideways}
    \includegraphics[width=.3211\linewidth]{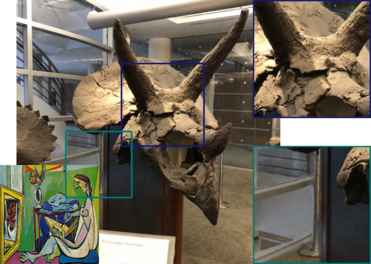}
    \includegraphics[width=.3069\linewidth]{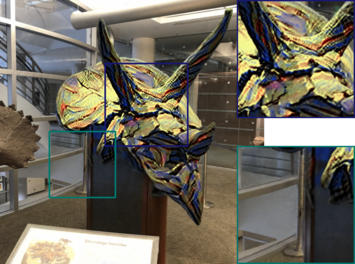}
    \includegraphics[width=.3069\linewidth]{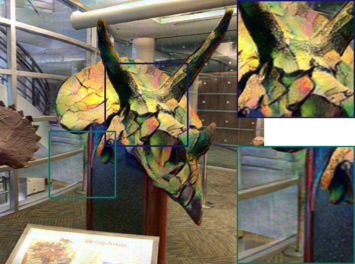} \\
    \caption{Qualitative comparison of our method to {\refnpr}.
    The top-row detailed view shows the selected region for stylization, whereas the bottom-row view shows a background region.
    Our method produces a more detailed stylization while minimizing background artefacts.}
    \label{fig:npr:qualitative}
\end{figure}

\paragraph{Quantitative Evaluation.}
For our quantitative evaluation, we measure MSE in the background with respect to the ground truth test set images.
For NeRF-Synthetic~\cite{Mildenhall2020NeRF}, we generate masks for the region to stylize using our method and use segmentation masks from ICE-NeRF~\cite{Lee2023ICENeRF} for LLFF~\cite{Mildenhall2020NeRF} scenes.
We report per-dataset results in \cref{tab:npr:quantitative:mse}.
In contrast to {\refnpr}, our approach demonstrates considerably reduced error rates, especially in synthetic scenes characterized by numerous occlusions. 
For forward-facing scenes, {\refnpr} benefits from less variation between camera poses but still generates about $3\times$ more errors than ours.

\begin{table}[h!]
  \centering
  \small
    \begin{tabular}{@{}lll@{}}\toprule
    Dataset & {\refnpr} & {\ours}  \\
    \midrule
    NeRF-Synthetic~\cite{Mildenhall2020NeRF} & 0.0466 & \textbf{0.0071}  \\
    LLFF~\cite{mildenhall2019llff} & 0.0073 & \textbf{0.0025}  \\
    \midrule
    Average & {0.0270} & \textbf{0.0048} \\
    \bottomrule
    \end{tabular}
  \caption{
  MSE ($\downarrow$) in the background with respect to the ground truth test images for our method and {\refnpr}~\cite{zhang2023refnpr}.
  {\ours} significantly outperforms {\refnpr} for synthetic and forward-facing scenes.
  }
  \label{tab:npr:quantitative:mse}
\end{table}

\begin{figure*}[ht!]
    \centering
    % Hotdog
    \begin{sideways}
        \hspace{2.1em}
        \emph{Hotdog}
    \end{sideways}
    \includegraphics[width=.24\linewidth]{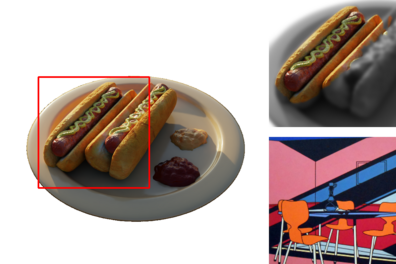}
    \includegraphics[width=.24\linewidth]{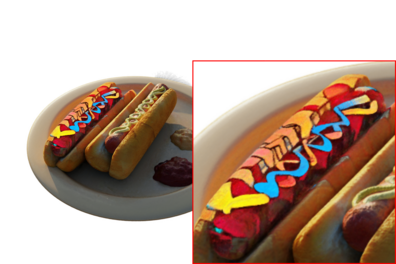}
    \includegraphics[width=.24\linewidth]{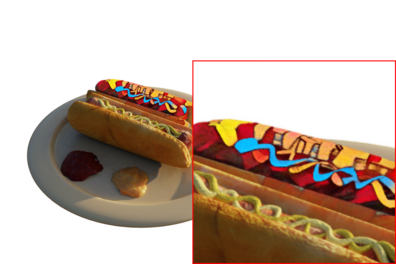}
    \includegraphics[width=.24\linewidth]{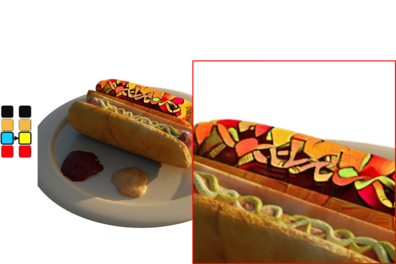}
    % Hotdog
    \begin{sideways}
        \hspace{2.3em}
        \emph{Chair}
    \end{sideways}
    \includegraphics[width=.24\linewidth]{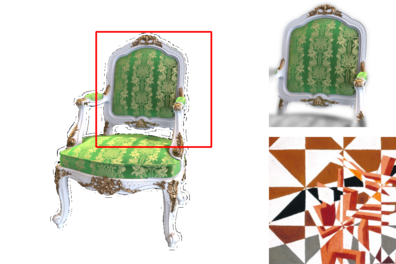}
    \includegraphics[width=.24\linewidth]{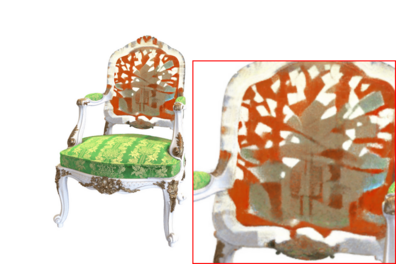}
    \includegraphics[width=.24\linewidth]{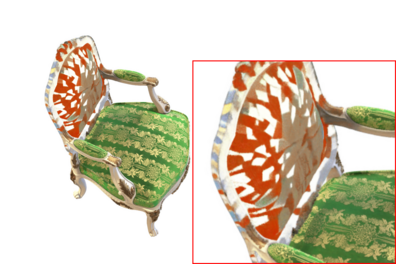}
    \includegraphics[width=.24\linewidth]{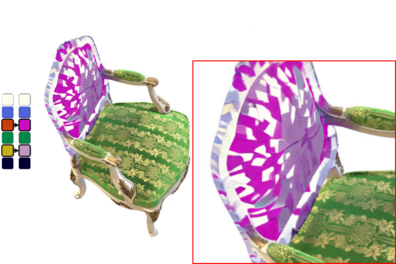}
    % Flower
    \begin{sideways}
        \hspace{1.5em}
        \emph{Flower}
    \end{sideways}
    \includegraphics[width=.24\linewidth]{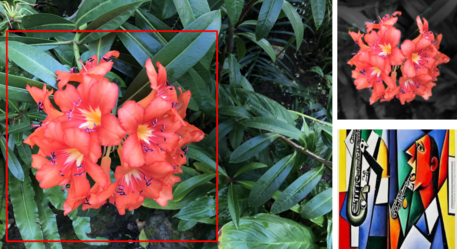}
    \includegraphics[width=.24\linewidth]{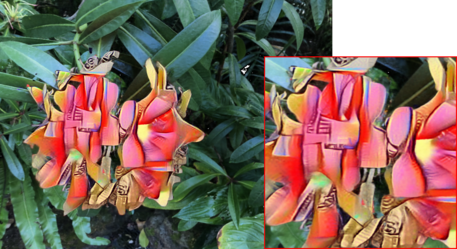}
    \includegraphics[width=.24\linewidth]{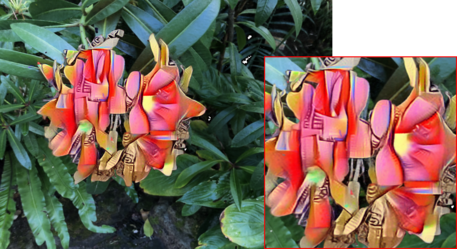}
    \includegraphics[width=.24\linewidth]{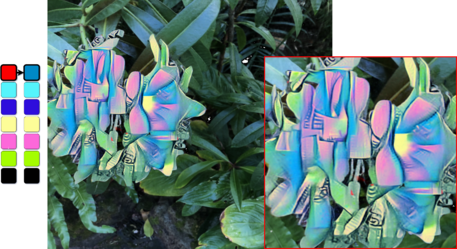}
    % Horns
    \begin{sideways}
        \hspace{1.7em}
        \emph{Horns}
    \end{sideways}
    \includegraphics[width=.24\linewidth]{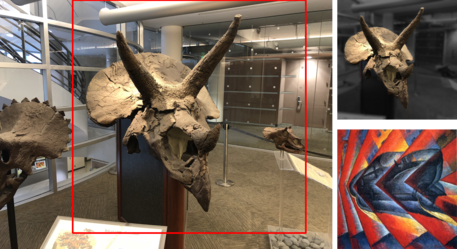}
    \includegraphics[width=.24\linewidth]{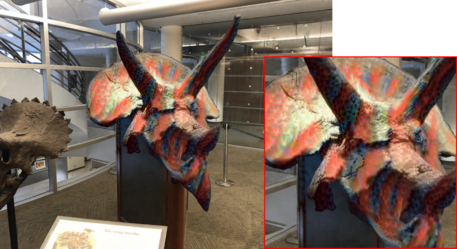}
    \includegraphics[width=.24\linewidth]{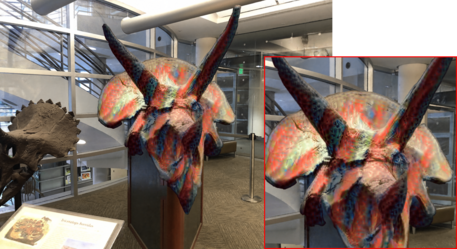}
    \includegraphics[width=.24\linewidth]{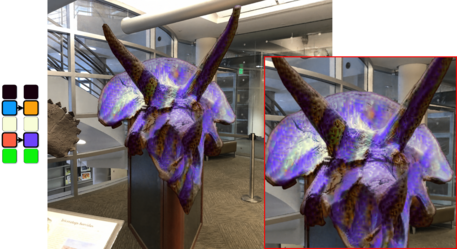}
    % Kitchen
    \begin{sideways}
        \hspace{1.2em}
        \emph{Kitchen}
    \end{sideways}
    \includegraphics[width=.24\linewidth]{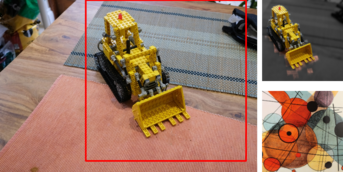}
    \includegraphics[width=.24\linewidth]{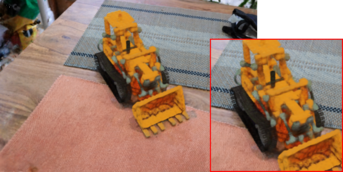}
    \includegraphics[width=.24\linewidth]{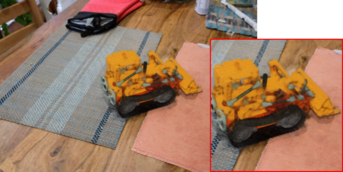}
    \includegraphics[width=.24\linewidth]{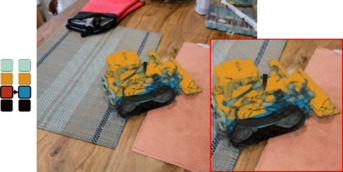}
    % Bonsai
    \begin{sideways}
        \hspace{1.2em}
        \emph{Bonsai}
    \end{sideways}
    \includegraphics[width=.24\linewidth]{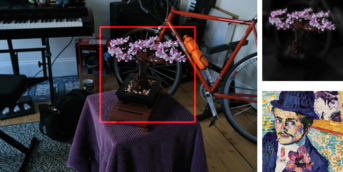}
    \includegraphics[width=.24\linewidth]{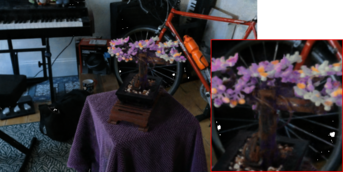}
    \includegraphics[width=.24\linewidth]{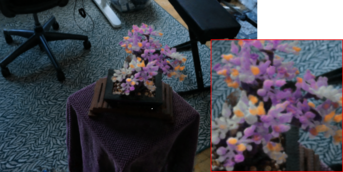}
    \includegraphics[width=.24\linewidth]{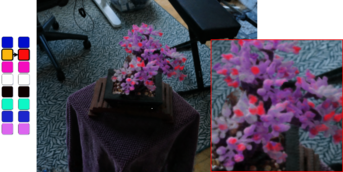}
    \caption{Results for local stylization for various scene types.
    {\ours} can faithfully transfer the style of an arbitrary style image to a selected region whilst minimizing errors in the background.
    Due to our decomposition into base colors, stylized regions remain editable.}
    \label{fig:stylization_results}
\end{figure*}

\paragraph{Qualitative Evaluation.}
In \cref{fig:npr:qualitative}, we show results for our method and {\refnpr} for synthetic and real-world scenes.
Our approach introduces fewer background artefacts compared to {\refnpr}, while stylizing the selected region with more detail.
In \cref{fig:stylization_results}, we show additional results on local, recolorable stylization for all scene types, including unbounded, real-world scenes~\cite{Barron2022MipNeRF360}.
As can be seen, our approach is compatible with diverse datasets, stylizes the selected region faithfully and produces intuitive color palettes for interactive recoloring.
\begin{figure}[ht!]
    \centering
    \includegraphics[width=\linewidth]{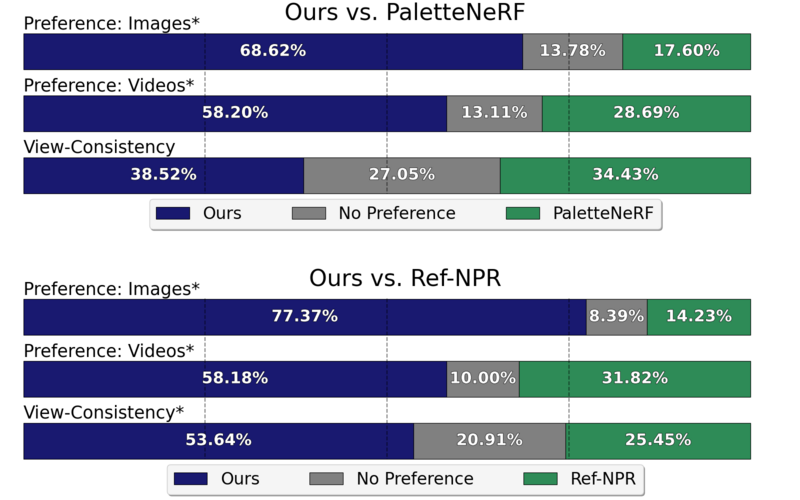}
    \caption{User study results.
    Participants prefer our method to related work for image and video outputs.
    Our method is also rated higher for view-consistency than {\refnpr}~\cite{zhang2023refnpr}.
    $^*$ indicates a statistical significance according to Wilcoxon signed rank tests.}
    \label{fig:userstudy}
\end{figure}

\subsection{User Study}
\label{sec:experiments:userstudy}
To further evaluate our approach, we conducted a user study comparing our approach to {\pnf}~\cite{kuang2022palettenerf} and {\refnpr}~\cite{zhang2023refnpr}.
We showed the participants pairs of recolored/stylized images with recoloring/style and target region inset and asked which result they preferred (a, b, or no-preference).
Additionally, we presented pairs of recolored/stylized videos of the same scenes and inquired about preference for visual quality and view-consistency (without reference).
We collected 847 responses from 31 participants as summarized in \cref{fig:userstudy}.
Participants prefer our approach for both images and videos and rated our approach as more view-consistent than {\refnpr}.
We refer to the supplementary material for details.

\subsection{Time Comparisons}
\label{sec:experiments:timcomparison}
We use the \emph{Flower} scene of the LLFF dataset~\cite{mildenhall2019llff}, for an exemplary time comparison:
With a pre-trained radiance field, {\pnf}~\cite{kuang2022palettenerf} requires $13.5$ min for recoloring a selected object, whereas our method accomplishes the same task in $3$ min.
When provided with stylized reference views, {\refnpr}~\cite{zhang2023refnpr} achieves scene stylization in $2.5$ min, whereas our approach takes $2$ min.
More importantly, we always provide previews after $\sim 20$s.
\section{Limitations}
\label{sec:limitations}
Although {\ours} is a flexible method for local appearance edits of NeRF, some challenges remain.
Similar to~\cite{Pang2023LocallyStylized}, our ability to modify stylized regions is restricted, specifically to adjusting palette bases post-training.
We leverage the pre-trained NeRF to perform geometry-aware appearance modifications.
As noted in other works~\cite{Barron2022MipNeRF360, NeffDONeRF, philipp2023floaters}, radiance fields often trade geometric fidelity for visual quality by modelling non-Lambertian effects with additional samples behind the surface, posing a challenge to our point-based optimization scheme. 
Particularly for real-world scenes, this may lead to reduced quality in the edited region.
Another disadvantage of {\ours} lies in the separation of optimization and distillation.
This design choice allows for interactive recoloring of stylized content as an intermediate step but incurs additional time for generating a modified training dataset and NeRF fine-tuning.
\section{Conclusion}
\label{sec:conclusion}
We present {\ours}, a unified framework for photorealistic and non-photorealistic appearance editing of NeRF.
By elegantly combining a palette-based decomposition with perceptual losses, we enable interactive recoloring of stylized regions.
We demonstrate state-of-the-art local appearance editing results, benefiting from our geometry-aware stylization in 3D.
{\ours} outperforms existing works quantitatively and qualitatively for local recoloring and local stylization.
By open-sourcing our approach we will bring NeRF-editing to a large audience.
{
    \small
    \bibliographystyle{ieeenat_fullname}
    \bibliography{main}
}
\appendix
\clearpage
\maketitlesupplementary
In this supplementary document, we provide more implementation details and quantitative results.
Additionally, we describe our GUI in detail and provide additional details for our user study.
\section{Additional Implementation Details}
In this section, we present more information on the network architecture of {\ours}.

\begin{figure}[ht!]
    \centering
    \includegraphics[width=\linewidth]{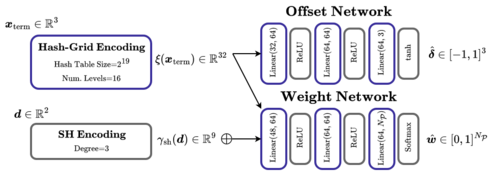}
    \caption{\textbf{Network architecture} for {\ours}.
    Blue components indicate trainable parameters and the input for the weight network is padded with ones.
    }
    \label{fig:sup:architecture}
\end{figure}

\subsection{Network Architecture}
In~\cref{fig:sup:architecture}, we show the network architecture for {\ours}.
As discussed in our main material, our module is NeRF-like: 
The sizes of the hash grid and linear layers as well as the spherical harmonics encoding are used similarly in {\ngp}~\cite{mueller2022instant}.
However, instead of two subsequent MLPs predicting density and color, we split our model and evaluate the weight and the offset network concurrently.

\subsection{Removal of Base Color Palettes}
As indicated in the main material, we start with $\numpalet=8$ base colors, which are initialized according to a uniform distribution, i.e. $\palet \sim \mathcal{U}[0,1)$.
Because $8$ color palettes are usually only required when we want to obtain smooth transitions during stylization, we remove color palettes which do not contribute significantly before the final $1500$ iterations.
To do this, we sample $10$ poses from the training dataset and evaluate our weight network.
We derive a per-palette mean contribution, which is always between $0$ and $1$.
Finally, we choose a threshold of $0.025$ and remove all color palettes which contribute less, which we reflect by updating our UI.

\subsection{Geometry-Aware Stylization}
To perform stylization which respects the learned geometry of our pre-trained NeRF, we perform depth estimation. 
As we set the depth to $0$ for all rays which did not intersect $\editgrid$, we would get incorrect results when computing our depth guidance image. 
Additionally, direct neighbors of rays which did not intersect the edit grid often do not accumulate full alpha inside $\editgrid$, leading to incorrect predictions for $\depth$.
To remedy the aforementioned issues, we multiply $\depthguide$ with the accumulated alpha inside the edit grid $\Talpha$, for both pixels involved in the computation and their direct neighbors:
\begin{align}
    (\depthguide)_{i, j} = (\depthguide)_{i, j} \cdot \left( \prod_{y=j-1}^{j+2} (\Talpha)_{i, y} ,\ \prod_{z=i-1}^{i+2} (\Talpha)_{z, j} \right)\label{eq:method:depthguide_fix}.
\end{align}

\subsection{Modifications to our Framework}
We include the following modifications to torch-ngp~\cite{torch-ngp} to improve reconstruction for real-world scenes, following {\pnf}~\cite{kuang2022palettenerf}.
By default, the background color is assumed to be white for real-world scenes --- iNGP might cheat with this static background color by not accumulating full $\alpha$ along a ray.  
We use a random background color during training, which leads to better depth estimates and a sparser occupancy grid.
In addition, we mark grid cells between each camera and its corresponding near plane as non-trainable.

\section{Ablation Studies}
\label{sec:experiments:ablation}
In this section, we provide additional analysis for the individual components of {\ours}.
In particular, we focus on the proposed losses for stylization and regularization of the color palette.

\subsection{Color Palette Regularization}
\begin{figure}[ht!]
    \centering
    \includegraphics[width=\linewidth]{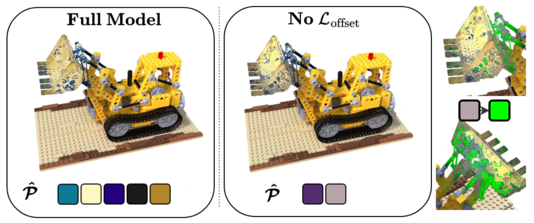}
    \caption{\textbf{Ablation study} on the effect of $\loffset$.
    Without this loss term, our model suffers from incorrect palette reconstruction and highly non-intuitive recoloring.}
    \label{fig:sup:ablation:recoloring}
\end{figure}

To measure the fidelity of the learned color palette $\palet$ quantitatively, we adopt the metrics from {\pnf}~\cite{kuang2022palettenerf} and measure sparsity with
\begin{equation}
    \mathcal{L}_{\text{sp}} = \frac{\sum_{i=1}^{\numpalet}\what_i}{\sum_{i=1}^{\numpalet}\what_i^2} - 1.
\end{equation}
In addition, we measure the total variation of the per-palette weight images $\what_i$. 
We present the metrics in \cref{tab:ablation:palettes}, where we recolored/stylized the shovel of the bulldozer in the \emph{Lego} scene.
Not using $\loffset$ causes extreme solutions, evidenced by $\mathcal{L}_{\text{sp}}$ and the non-intuitive recoloring in~\cref{fig:sup:ablation:recoloring}.
Not using $\lweight$ extracts intuitive palettes but uses more base colors and results in high values for $\mathcal{L}_{\text{sp}}$.
{\ours} achieves the best sparsity and requires the fewest base colors.

\begin{table}[ht!]
  \small
  \centering
    \begin{tabular}{@{}lllllll@{}}\toprule
    \multirow{3}{*}{Method} & \multicolumn{3}{c}{Stylization} & \multicolumn{3}{c}{Recoloring}\\
    \cmidrule(lr){2-4}\cmidrule(lr){5-7}
     & $\numpalet$ & $\mathcal{L}_{\text{sp}} \downarrow$ & TV $\downarrow$ & $\numpalet$ & $\mathcal{L}_{\text{sp}} \downarrow$ & TV $\downarrow$ \\
    \midrule
    w/o $\loffset$ & 8  & 6.046         & \bf{0.005}     & 8 & 6.304 & \textbf{0.005}\\
    w/o $\lweight$ & 8  & {3.033}       & {0.026}        & 8 & 2.375 & {0.042}\\
    \midrule
    {\ours}        & \textbf{6}  & \bf{2.129}    & {0.029}      & \textbf{5} & \textbf{0.859} & {0.107} \\
    \bottomrule
    \end{tabular}
  \caption{{Quantitative Evaluation} of our proposed color palette regularization.
  Our full model achieves the best metrics for sparsity and uses the fewest base colors.}
  \label{tab:ablation:palettes}
\end{table}

\subsection{Geometry-Aware Stylization Losses}
\begin{figure}[ht!]
    \centering
    \includegraphics[width=\linewidth]{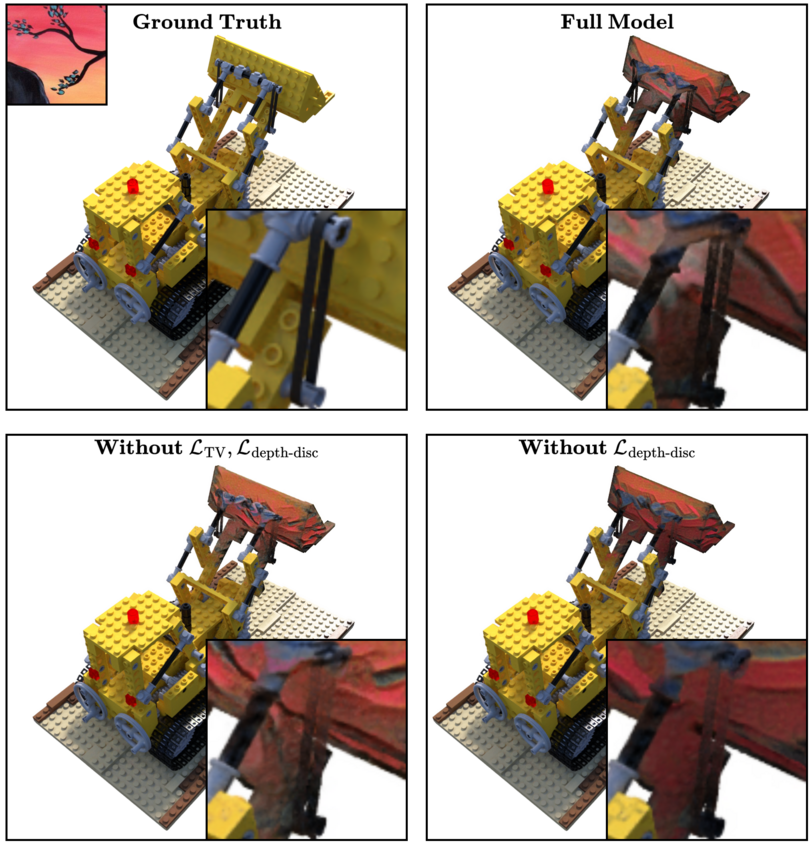}
    \caption{\textbf{Ablation study} on the effectiveness of our proposed losses for 3D-aware stylization.
    Our full model retains the most detail during stylization.}
    \label{fig:sup:ablation:details}
\end{figure}

Na\"ively applying 2D style transfer losses often results in removal of smaller structure, \eg the black rubber bands in the \emph{Lego} scene disappear to obtain a more coherent stylization.
We demonstrate the effectiveness of our proposed counter-measures in~\cref{fig:sup:ablation:details}.
As can be seen, {\ours} retains small structures well during stylization compared to only using $\ltv$ and the na\"ive approach without any geometry-conditioned losses.
When stylizing, we additionally train {\ours} without $\lstyle, \loffset, \ltv$ for the first 1000 iterations, resulting in a well-initialized palette $\palet$.

\subsection{View-Consistency}
In addition to our experiments in the main paper, we evaluate a view-consistency metric from SNeRF~\cite{nguyen2022snerf}, which is a generalization of Lai~\etal~\cite{Lai_2018_ECCV} to the NeRF setting.
Following SNeRF, we estimate the optical flow using RAFT~\cite{teed2020raft} and use the occlusion detection method from Ruder~\etal~\cite{ruder2016artistic} to derive an occlusion mask $O$.
Importantly, we compute the optical flow on a video sequence rendered from a pre-trained NeRF.
Then, we compute the MSE between the warped view $\bar{\vec{V}}_{i+\delta}$ and the rendered view ${\vec{V}}_{i+\delta}$ using
\begin{equation}
    E_{\text{warp}}({\vec{V}}_{i+\delta}, \bar{\vec{V}}_{i+\delta}) = \frac{1}{|O|} \left\|\bar{\vec{V}}_{i+\delta} - {\vec{V}}_{i+\delta}\right\|_2^2.
\end{equation}
In addition, we also measure LPIPS~\cite{zhang2018unreasonable} between $\bar{\vec{V}}_{i+\delta}$ and ${\vec{V}}_{i+\delta}$, following StyleRF~\cite{liu2023stylerf}.
We show our results in~\cref{tab:sup:view-cons}, where we analyzed $6$ video sequences for stylization and recoloring.
We evaluate short-range consistency ($\delta=1$) and long-range consistency ($\delta=7$).
{\ours} outperforms {\refnpr}~\cite{zhang2023refnpr} for all metrics.
{\pnf}~\cite{kuang2022palettenerf} with semantic features achieves slightly better results for MSE, but performs worse for LPIPS.
These quantitative results align with the results for our user study. 

\begin{table}[ht!]
  \setlength{\tabcolsep}{5pt}
  \small
    \begin{tabular}{@{}llllll@{}}\toprule
        &\multicolumn{5}{c}{\emph{Stylization Consistency}} \\
                \cmidrule{2-6}
        &\multicolumn{2}{c}{ {Short-Range}} && \multicolumn{2}{c}{ {Long-Range}} \\
        \cmidrule{2-3}\cmidrule{5-6}
        &MSE ($\downarrow$) & LPIPS ($\downarrow$) && MSE ($\downarrow$) & LPIPS ($\downarrow$) \\
        \midrule
        \ours &\textbf{0.0252} & \textbf{0.0650} && \textbf{0.1932} & \textbf{0.2253} \\
        \refnpr & {0.0264} & 0.0722 && {0.1973} & 0.2482 \\\bottomrule
        &&&&& \\[-.9ex]
        &\multicolumn{5}{c}{\emph{Recoloring Consistency}} \\
        \cmidrule{2-6}
        &\multicolumn{2}{c}{ {Short-Range}} && \multicolumn{2}{c}{ {Long-Range}} \\
        \cmidrule{2-3}\cmidrule{5-6}
        &MSE ($\downarrow$) & LPIPS ($\downarrow$) && MSE ($\downarrow$) & LPIPS ($\downarrow$) \\
        \midrule
        \ours &{0.0167} & \textbf{0.0587} && {0.0934} & \textbf{0.2063} \\
        \pnf & \textbf{0.0164} & 0.0634 && \textbf{0.0925} & 0.2080 \\[-.7ex]
        {\scriptsize (semantic)}& & && & \\[-.7ex]
        \bottomrule
    \end{tabular}
  \caption{\textbf{View-Consistency comparison} of our method, {\refnpr}~\cite{zhang2023refnpr} and {\pnf}~\cite{kuang2022palettenerf} with semantic guidance.
  We measure short-range consistency ($\delta=1$) and long-range consistency ($\delta=7$).}
  \label{tab:sup:view-cons}
  \centering
\end{table}

\section{Additional User Study Details}
We conduct our user study for a comparison with state-of-the-art methods for local recoloring and local style transfer in scenes represented by NeRFs.
We conduct the study on an iPad 9\textsuperscript{th} Gen with a 10.2" display.
All images and videos used for our user study are included in the supplementary material.

\paragraph{Images: Local Recoloring.}
For local recoloring, we select scenes from the LLFF dataset~\cite{mildenhall2019llff} (\emph{Flower, Horns, Fortress, Orchids, Trex}) and the {\threesixty} dataset~\cite{Barron2022MipNeRF360} (\emph{Kitchen, Bonsai, Room}) and compare against {\pnf}~\cite{kuang2022palettenerf} with semantic guidance.
For the per-image comparisons, we prepare 11 different recolorings, where we tried to align the results as much as possible for a fair comparison.
For each pair of images (one from our method, one from {\pnf}), the user is shown the reference test set image, the region we want to recolor and a color change.
We randomize the order in which we present the different conditions.
Users were instructed to choose the image they prefer (a, b, or no-preference) based on background artefacts, image quality and personal preference.

\begin{figure*}[ht!]
    \centering
    \includegraphics[width=.95\linewidth]{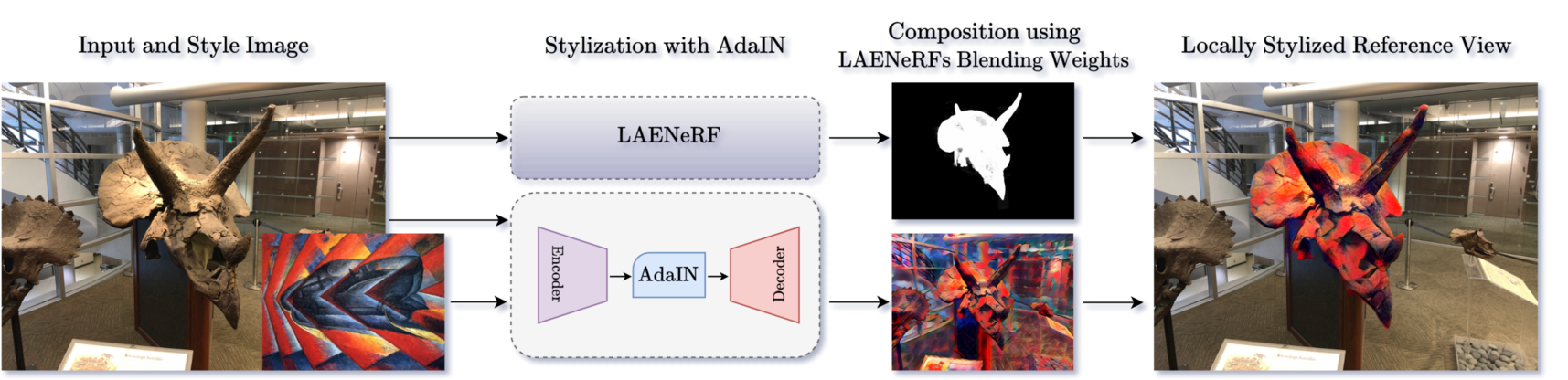}
    \caption{\textbf{Generation of locally stylized reference images} for {\refnpr}~\cite{zhang2023refnpr} using AdaIN~\cite{huang2017arbitrary} and {\ours}'s blending weights.}
    \label{fig:npr:style}
\end{figure*}

\paragraph{Images: Local Stylization.}
For local stylization, we select scenes from the LLFF dataset~\cite{mildenhall2019llff} (\emph{Flower}, \emph{Horns}, \emph{Trex}) and the NeRF-Synthetic dataset~\cite{Mildenhall2020NeRF} (\emph{Lego, Hotdog, Chair, Drums}) and compare against {\refnpr}~\cite{zhang2023refnpr}.
For the per-image comparison, we prepare 9 different stylizations, where we tried to align the results as much as possible for a fair comparison.
Instead of showing the user a color change, we now show the user the style image $\styleimg$.
The images for {\refnpr} were generated leveraging AdaIN~\cite{huang2017arbitrary}, as shown in \cref{fig:npr:style}.
The testing modality is identical to local recoloring, however, users were also instructed to consider the transfer of style from the given style image $\styleimg$ to the selected region.

\paragraph{Videos.}
In addition to images, users were also shown videos of recolorings and stylizations.
For this modality, we did not provide a specified region, target color or style images.
Users were instead instructed to select their preferred video based on (1) visual appearance and (2) view-consistency.

\paragraph{Participant Details.}
Of the $31$ participants of our user study, $58\%$ had at least some prior experience with NeRFs.
In addition, $29\%$ used a visual aid during the user study.

\paragraph{Statistical Evaluation.}
We compute the Wilcoxon signed rank test to determine whether there is statistical significance in the preference for one or the other method. 
All preference scores indicate a significant preference for {\ours} with recoloring ($Z=8850.0;\ p<0.0001$), stylization ($Z=5000.0;\ p<0.0001$), video recoloring ($Z=1872.5;\ p<0.0005$), and video stylization ($Z=1750.0;\ p<0.005$). 
The view consistency scores indicate a preference for our method for stylization ($Z=1232.0;\ p<0.001$), but no significant difference for recoloring ($Z=1890.0;\ p>0.5$). 

\section{GUI}
{\ours} incorporates a GUI building on~\cite{jambon2023nerfshop, torch-ngp} for real-time, interactive appearance editing of NeRFs.
Here, we present several key components of our user interface.

\begin{figure}[ht!]
    \centering
    \includegraphics[width=\linewidth]{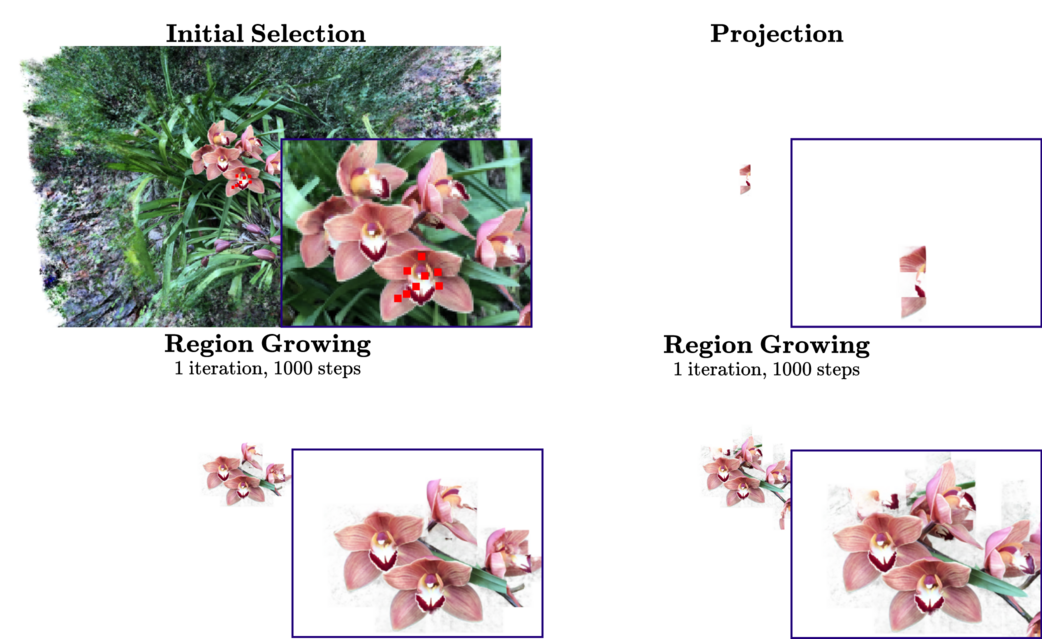}
    \caption{\textbf{Region selection} with our interactive GUI.
    The user first clicks on the screen (shown as red squares). 
    For each selected ray, we compute the estimated ray termination and map it to the nearest cell in $\editgrid$.
    Finally, this initial selection can be extended with region growing (2 iterations shown here).}
    \label{fig:sup:gui:selection}
\end{figure}

\subsection{Region Selection and Growing}
We provide an intuitive region selection process, based on NeRFShop~\cite{jambon2023nerfshop}.
First, the user scribbles on the image rendered from the current camera pose in our real-time viewer.
For each selected ray, we perform raymarching and map the estimated ray termination to a the nearest voxel, which we set in $\editgrid$, representing our initial selection.
Next, the initial selection can be extended with region growing, where the user controls growing by the number of voxels which we pop from our growing queue per-iteration and the total number of iterations.
Our GUI shows the selection after each region growing step.
We can optionally create a growing grid $\growgrid$ from our selection to model smooth transitions.
Further, we incorporate binary operations with a second grid for more intuitive addition and deletion of selected voxels.
Due to the region growing procedure and easy deletion of voxels with a second grid and binary operations, our method is quite insensitive to the user scribble provided as input.
The complete workflow can be seen in~\cref{fig:sup:gui:selection}.

\subsection{Style Image Selection}
\begin{figure}[ht!]
    \centering
    \includegraphics[width=.32\linewidth]{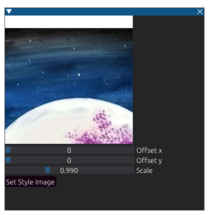}
    \includegraphics[width=.32\linewidth]{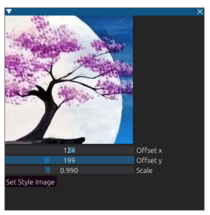}
    \includegraphics[width=.32\linewidth]{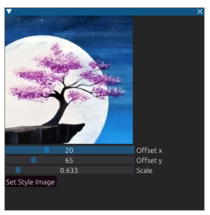}
    \caption{\textbf{GUI interface} for style image selection.}
    \label{fig:sup:gui:styleimg}
\end{figure}
Our {\ours} module requires style images $\styleimg \in \R^{256 \times 256 \times 3}$ for stylization.
To enhance usability of our approach and to support arbitrary style images, we provide an intuitive GUI for style image selection.
As can be seen in~\cref{fig:sup:gui:styleimg}, our GUI supports zooming and cropping arbitrary images to the required size.

\subsection{More Palette Control}
\begin{figure}[ht!]
    \centering
    \includegraphics[width=\linewidth]{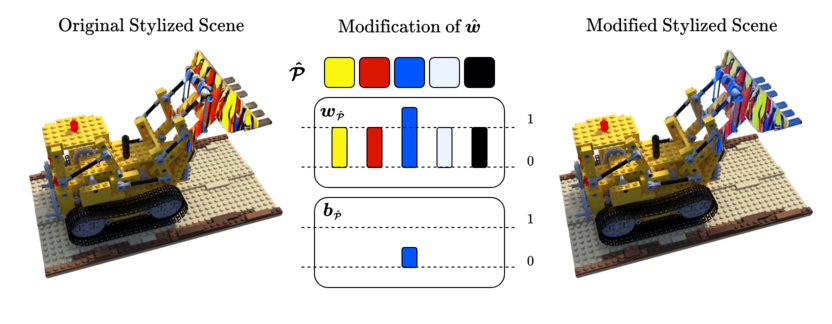}
    \caption{\textbf{Demonstration of user-guided modifications to} $\what$.
    We enable modification of stylized regions by changing the importance of individual base colors $\palet_i$ via modification of $\paletweight, \paletbias$.}
    \label{fig:sup:paletmod}
\end{figure}

To provide users with more control over the stylized/recolored region, we enable a linear transformation of the learned weights $\what$.
To this end, we introduce palette weights $\paletweight \in \R^{\numpalet}$ and palette biases $\paletbias \in [-1, 1]^{\numpalet}$, which transform the weights according to
\begin{align}
    \begin{split}
        \what &= \min(\what \cdot \paletweight + \paletbias, 0), \\
        \what &= \frac{\what}{\vec{1}^T \what}.
    \end{split}
\end{align}
We initialize $\paletweight = \vec{1},\ \paletbias = \vec{0}$ and let the user guide these parameters after {\ours} is fully-trained, as can be seen in~\cref{fig:sup:paletmod}.

\subsection{Preview Mode \& GUI demonstration}
To enable interactive recoloring and stylization, our approach supports real-time rendering during training of our {\ours} module.
After the training dataset has been extracted, which takes $\sim 15$ seconds depending on the dataset, users can watch our neural module converge.
As we only handle views of the training dataset during optimization, this efficient dataset extraction results in fast convergence.

To render novel views during training, enabling interactive previews, we query $\nerf$ to obtain the estimated ray termination $\xterm$ and $\Talpha$, as we do during training dataset extraction.
If $\Talpha > 0.5$, we use $\xterm$ as input to {\ours} and output the rendered image.
Consequently, due to the fast stylization process, users have the ability to stop training after few seconds and choose another style image if they find the result unappealing.

\subsection{Detailed Time Comparisons}
We provide a more detailed analysis of the time comparison to {\pnf}~\cite{kuang2022palettenerf} in~\cref{tab:sup:time}.
As can be seen, {\ours} performs the recoloring task much faster, although the same NeRF backbone is used.

\begin{table}[ht!]
  \centering
  \small
    \begin{tabular}{@{}cccccc@{}}\toprule
    \multicolumn{4}{c}{{\pnf}} && \\
    \cmidrule{1-4}
    {\scriptsize Semantic} &  {\scriptsize Train} &  {\scriptsize Extract} &  {\scriptsize Train} & & Total \\[-.9ex]
    {\scriptsize Features} &  {\scriptsize NeRF}  & {\scriptsize Palette}  & {\scriptsize {\pnf}} && \\[-.7ex]
    \midrule
    $90$s & $153$s & $10$s & $572$s & & $815$s \\
    \midrule
    \multicolumn{5}{c}{{\ours}} &\\
    \cmidrule{1-5}
        {\scriptsize Train} &  {\scriptsize Edit} &  {\scriptsize Train} &  {\scriptsize Distill} & {\scriptsize Fine-Tune} & Total \\[-.9ex]
    {\scriptsize NeRF} &  {\scriptsize Dataset}  & {\scriptsize {\ours}}  & {\scriptsize Dataset} & {\scriptsize NeRF} & \\[-.7ex]
        \midrule
    $153$s & $16$s & $133$s & $32$s & $129$s & $\mathbf{463}$\textbf{s} \\
    \bottomrule
    \end{tabular}
  \caption{\textbf{Detailed timing comparison} of our method and {\pnf}~\cite{kuang2022palettenerf}.}
  \label{tab:sup:time}
\end{table}

\section{Per Scene Results for Quantitative Evaluation}
For a better reasoning behind the quantitative results in the main paper and to facilitate more research in the area of local appearance editing, we report per scene results for recoloring and stylization.

\subsection{LLFF Recoloring}
\begin{figure}
    \centering
         \begin{minipage}[t]{0.3\linewidth}
        \centering
        \footnotesize
        \text{{\ours}}
    \end{minipage}
     \begin{minipage}[t]{0.3\linewidth}
        \centering
        \footnotesize
        \text{{\pnf}~\cite{kuang2022palettenerf}}
    \end{minipage}
    \begin{minipage}[t]{0.3\linewidth}
        \centering
        {\footnotesize
        \text{{\pnf}~\cite{kuang2022palettenerf}}} \\[-1ex]
        {\scriptsize
        \text{(Semantic)}}
    \end{minipage} \\
    \includegraphics[width=.3\linewidth]{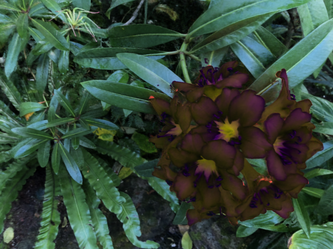}
    \includegraphics[width=.3\linewidth]{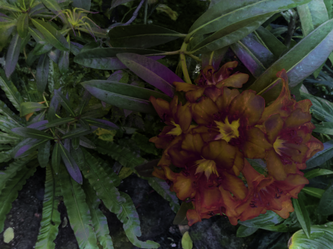}
    \includegraphics[width=.3\linewidth]{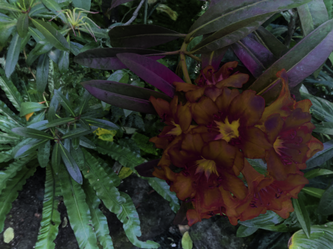}
    \caption{\textbf{Example of Color Edits} for the LLFF dataset~\cite{mildenhall2019llff} for our method and {\pnf}~\cite{kuang2022palettenerf}.}
    \label{fig:llff:recoloring:idek}
\end{figure}

For the LLFF dataset~\cite{mildenhall2019llff}, we show our per-recoloring results in~\cref{tab:sup:llff:recoloring}.
All masks were provided to us by the authors of {\icenerf}~\cite{Lee2023ICENeRF}.
For our color edits, we also include substantial recolorings, as can be seen in~\cref{fig:llff:recoloring:idek}.
{\pnf}~\cite{kuang2022palettenerf} requires changing all color palettes for this specific example due to their decomposition, which introduces background artefacts, even when semantic features are used for guidance.
Our approach geometrically segments the region in 3D, leading to good recoloring results without introducing significant artefacts in the background.

\subsection{{\threesixty} Recoloring}
\begin{figure}[ht!]
    \centering
    \includegraphics[width=\linewidth]{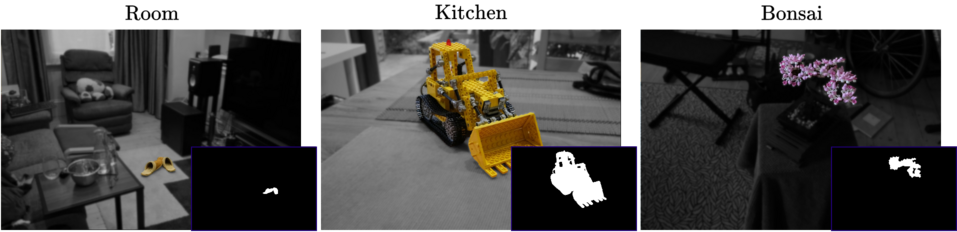}
    \caption{\textbf{Foreground masks} for the {\threesixty} dataset~\cite{Barron2022MipNeRF360}.
    We lowered brightness and saturation and additionally blurred the background using the masks.}
    \label{fig:sup:360:masks}
\end{figure}

For the {\threesixty} dataset~\cite{Barron2022MipNeRF360}, we report per-recoloring results in~\cref{tab:sup:360:recoloring}.
We measure MSE in the background of the selected region, with masks extracted using Segment Anything~\cite{kirillov2023segmentanything} describing the selected object.
We obtain $14$ masks for test scenes of the \emph{Bonsai} scene, $32$ masks for the \emph{Kitchen} scene and $18$ masks for the \emph{Room} scene.
We show some examples of the extracted masks in~\cref{fig:sup:360:masks}.

\subsection{Local Stylization}
For local stylization, we report per scene results in~\cref{tab:sup:mse:stylization}.
As can be seen, our method outperforms {\refnpr}~\cite{zhang2023refnpr} for every scene.

\begin{table}[h!]
  \centering
  \small
    \begin{tabular}{@{}lllll@{}}\toprule
    Method &\multicolumn{3}{c}{{NeRF-Synthetic~\cite{Mildenhall2020NeRF}}}&\\
    \cmidrule(lr){2-4}
     & \emph{Chair} & \emph{Drums} & \emph{Lego} & Average \\
    \midrule
    {\refnpr}~\cite{zhang2023refnpr} & 0.0357 & 0.0808 & 0.0233 & 0.0466\\
    {\ours} & 0.0037 & 0.0154 & 0.0021 & \textbf{0.0071}\\
    \midrule
    &\multicolumn{3}{c}{{LLFF~\cite{mildenhall2019llff}}}&\\
    \cmidrule(lr){2-4}
     & \emph{Horns} & \emph{Flower} & \emph{Trex} & Average \\
    \midrule
    {\refnpr}~\cite{zhang2023refnpr} & 0.0075 & 0.0079 & 0.0064 & 0.0073\\
    {\ours} & 0.0024 & 0.0026 & 0.0025 & \textbf{0.0025}\\
    \bottomrule
    \end{tabular}
  \caption{\textbf{MSE ($\downarrow$) in the background} per scene for local stylization for our method and {\refnpr}~\cite{zhang2023refnpr}.}
  \label{tab:sup:mse:stylization}
\end{table}

\begin{table*}
  \centering
  \small
    \begin{tabular}{@{}lllllllll@{}}\toprule
     & \multicolumn{7}{c}{\emph{Horns}} & \\
     \cmidrule{2-8}
     & {\scriptsize cyan}& {\scriptsize dark red}& {\scriptsize green}& {\scriptsize grey}& {\scriptsize magenta}& {\scriptsize light red} & {\scriptsize yellow} &{\scriptsize Average}\\\midrule
     {\pnf}&0.0026 &0.0437 &0.0138 &0.0347 &0.0004 &0.0237 &0.0173 &0.0195\\
     {\pnf} (semantic) &0.0008 &0.0084 &0.0011 &0.0072 &0.0004 &0.0016 &0.0001 &0.0028\\
     {\ours}&0.0010 &0.0010 &0.0010 &0.0010 &0.0008 &0.0011 &0.0007 & \textbf{0.0010}\\\midrule
     & \multicolumn{7}{c}{\emph{Fortress}} & \\
     \cmidrule{2-8}
     & {\scriptsize dark turquoise}& {\scriptsize magenta}& {\scriptsize dark orange}& {\scriptsize white}& {\scriptsize indigo}& {\scriptsize blueviolet} & {\scriptsize greenyellow} & {\scriptsize Average}\\\midrule
     {\pnf}& 0.0007 &0.0009 &0.0013 &0.0010 &0.0009 &0.0018 &0.0011 &0.0011\\
     {\pnf} (semantic)& 0.0001 &0.0001 &0.0001 &0.0001 &0.0004 &0.0002 &0.0002 & \textbf{0.0002}\\
     {\ours}&0.0002 &0.0003 &0.0002 &0.0002 &0.0002 &0.0002 &0.0002& \textbf{0.0002}\\\midrule
     & \multicolumn{7}{c}{\emph{Flower}} & \\
     \cmidrule{2-8}
     & {\scriptsize dark purple}& {\scriptsize light red}& {\scriptsize green}& {\scriptsize yellow}& {\scriptsize green blue}& {\scriptsize orange} & {\scriptsize purple} &{\scriptsize Average}\\\midrule
     {\pnf} & 0.0172 &0.0005 &0.0057 &0.0099 &0.0076 &0.0100 &0.0026 &0.0076\\
     {\pnf} (semantic)&0.0102 &0.0004 &0.0009 &0.0010 &0.0012 &0.0004 &0.0012 &0.0022\\
     {\ours}&0.0006 &0.0006 &0.0007 &0.0010 &0.0006 &0.0006 &0.0006 &\textbf{0.0007}\\
    \bottomrule
    \end{tabular}
  \caption{\textbf{MSE ($\downarrow$) in the background per recoloring} for the LLFF dataset~\cite{mildenhall2019llff} for our method and {\pnf}~\cite{kuang2022palettenerf} with and without semantic guidance.}
  \label{tab:sup:llff:recoloring}
\end{table*}

\begin{table*}
  \centering
  \small
    \begin{tabular}{@{}llllllll@{}}\toprule
     & \multicolumn{6}{c}{\emph{Room}, Sandles} & \\
     \cmidrule{2-7}
     & {\scriptsize blue}& {\scriptsize green}& {\scriptsize red}& {\scriptsize dark grey}& {\scriptsize purple}& {\scriptsize greenyellow} & {\scriptsize Average}\\\midrule
     {\pnf}&0.0384 &0.0186 &0.0046 &0.0339 &0.0291 &0.0050 &0.0216\\
     {\pnf} (semantic) & 0.0046 &0.0027 &0.0023 &0.0136 &0.0059 &0.0047 &0.0056\\
     {\ours}&0.0015 &0.0015 &0.0015 &0.0015 &0.0015 &0.0015 &\textbf{0.0015}\\\midrule
     & \multicolumn{6}{c}{\emph{Bonsai}, Flower} & \\
     \cmidrule{2-7}
     & {\scriptsize blueviolet}& {\scriptsize light yellow}& {\scriptsize red}& {\scriptsize lime}& {\scriptsize purple}& {\scriptsize black} & {\scriptsize Average}\\\midrule
     {\pnf} & 0.0025 &0.0034 &0.0040 &0.0038 &0.0028 &0.0048 &0.0036\\
     {\pnf} (semantic)&0.0015 &0.0015 &0.0015 &0.0016 &0.0016 &0.0016 &0.0016\\
     {\ours}&0.0011 &0.0011 &0.0011 &0.0011 &0.0011 &0.0012 &\textbf{0.0011}\\\midrule
     & \multicolumn{6}{c}{\emph{Kitchen}, Bulldozer} & \\
     \cmidrule{2-7}
     & {\scriptsize dark turquoise}& {\scriptsize magenta}& {\scriptsize dark orange}& {\scriptsize white}& {\scriptsize indigo}& {\scriptsize blueviolet} & {\scriptsize Average}\\\midrule
     {\pnf}& 0.0044 &0.0039 &0.0025 &0.0038 &0.0034 &0.0568 &0.0125\\
     {\pnf} (semantic)& 0.0028 &0.0028 &0.0024 &0.0027 &0.0027 &0.0027 &0.0027\\
     {\ours}&0.0023 &0.0022 &0.0021 &0.0021 &0.0021 &0.0021 &\textbf{0.0022}\\
    \bottomrule
    \end{tabular}
  \caption{\textbf{MSE ($\downarrow$) in the background per recoloring} for the {\threesixty} dataset~\cite{Barron2022MipNeRF360} for our method and {\pnf}~\cite{kuang2022palettenerf} with and without semantic guidance.
  {\ours} outperforms {\pnf} for every recoloring, demonstrating effectiveness for diverse recolorings.}
  \label{tab:sup:360:recoloring}
\end{table*}

\end{document}